\documentclass[lettersize,journal]{IEEEtran}
\usepackage{amsmath,amsfonts}
\usepackage{algorithmic}
\usepackage{array}
\usepackage[caption=false,font=normalsize,labelfont=sf,textfont=sf]{subfig}
\usepackage{textcomp}
\usepackage{stfloats}
\usepackage{url}
\usepackage{verbatim}
\usepackage{graphicx}
\usepackage{multirow}
\def\BibTeX{{\rm B\kern-.05em{\sc i\kern-.025em b}\kern-.08em
    T\kern-.1667em\lower.7ex\hbox{E}\kern-.125emX}}
\usepackage{balance}
\begin{document}
\title{Graph-Ensemble Learning Model for Multi-label Skin Lesion Classification using Dermoscopy and Clinical Images}

\author{Peng TANG,~\IEEEmembership{member,~IEEE,}, Yang NAN,~\IEEEmembership{member,~IEEE,}, Tobias Lasser ~\IEEEmembership{member,~IEEE,}
	
\thanks{Peng TANG and Tobias Lasser are with the Department of Informatic at Technical University of Munich, Garching, German, and Munich Institute of Biomedical Engineering.
	Yang NAN is currently a PhD student and research assistant in NHLI, Imperial College London.
		}}

\markboth{Journal of \LaTeX\ Class Files,~Vol.~18, No.~9, September~2020}%
{How to Use the IEEEtran \LaTeX \ Templates}

\maketitle

\begin{abstract}
Many skin lesion analysis (SLA) methods recently focused on developing a multi-modal-based multi-label classification method due to two factors. 
The first is multi-modal data, i.e., clinical and dermoscopy images, which can provide complementary information to obtain more accurate results than single-modal data. 
The second one is that multi-label classification, i.e., seven-point checklist (SPC) criteria as an auxiliary classification task can not only boost the diagnostic accuracy of melanoma in the deep learning (DL) pipeline but also provide more useful functions to the clinical doctor as it is commonly used in clinical dermatologist's diagnosis.
However, most methods only focus on designing a better module for multi-modal data fusion; few methods explore utilizing the label correlation between SPC and skin disease for performance improvement.
This study fills the gap that introduces a Graph Convolution Network (GCN) to exploit prior co-occurrence between each category as a correlation matrix into the DL model for the multi-label classification. 
However, directly applying GCN degraded the performances in our experiments; we attribute this to the weak generalization ability of GCN in the scenario of insufficient statistical samples of medical data.
We tackle this issue by proposing a Graph-Ensemble Learning Model (GELN) that views the prediction from GCN as complementary information of the predictions from the fusion model and adaptively fuses them
by a weighted averaging scheme, which can utilize the valuable information from GCN while avoiding its negative influences as much as possible.
To evaluate our method, we conduct experiments on public datasets.
The results illustrate that our GELN can consistently improve the classification performance on different datasets and that the proposed method can achieve state-of-the-art performance in SPC and diagnosis classification.
\end{abstract}


\section{Introduction}
\IEEEPARstart{S}{kin} disease is one of the most common reasons for people to visit clinics \cite{Schofield}, and melanoma is the most dangerous type of skin cancer \cite{Garbe, Ward}.
In 2023, according to the report \cite{ACS2023}, about 100 thousand melanoma patients are estimated to be diagnosed in the United States, and about 8000 will die due to the disease.
Early diagnosis can remarkably increase the five-year survival rate from 32$\%$ of the distant stage to 99$\%$ of the local stage \cite{ACS2023}, while the number of experienced dermatologists limits the prevention of melanoma in the early stage. 
So, developing an automated skin lesion analysis (SLA) method is expected to improve the doctor's efficiency and solve this problem.
In the clinics, the routine diagnosis of skin diseases is firstly examined by the clinical image using a digital camera, then dermoscopy \cite{Fu2022, Herschorn2012, Henning2008}.
Clinical images provide the lesion's information of color and geometry, and dermoscopy images show the magnified details of the lesion, such as the vascularity \cite{Yang2018, Bi2019}.
So both modalities images are considered together for the final assessment of the lesion based on the prior diagnostic rules, such as the Seven-Point Checklist (SPC) rule \cite{Fu2022}. 
The SPC is widely-used diagnosis criteria that examine the seven types of visual characteristics of skin lesions, i.e., vascular structures (VS), blue whitish veil (BWV), dots and globules (DaG), regression 
structures (RS), streaks (STR) and pigment network (PIG) \cite{Herschorn2012, Argenziano2001, Argenziano2011}. Each visual characteristic will be given a score,and if the total score is greater than the given threshold, the corresponding lesion would most likely be considered melanoma and be biopsied \cite{Fu2022}.

Traditional methods \cite{Abbas2013,XieF2016,Celebi2008, Rajpara2009, Fabbrocini2014} focused on designing the representative hand-crafted features of the skin lesion and then input them into machine learning classifiers to obtain the diagnosis and SPC result \cite{Rajpara2009}. However, these methods are not well-generalized because of the unreliable features.
Convolutional neural network (CNN)-based methods \cite{Yu2017, Zhang2019, Tang2020, Liu2022} significantly enhance the diagnostic accuracy compared to traditional methods.
However, these methods are only based on dermoscopy images while ignoring the integration of clinical images into the diagnostic pipeline.

More recently, many multi-modal methods \cite{Kawahara2019, Yap2018, Ge2017, Bi2020, Wang2022, Tang2022, Fu2022} have been proposed to use both clinical images and dermoscopy images for skin lesion classification.
Most of them \cite{Kawahara2019, Yap2018, Ge2017, Bi2020, Wang2022, Tang2022}, such as EmbeddingNet \cite{Yap2018}, HcCNN \cite{Bi2020}, FusionM4Net \cite{Tang2022}, concentrate on developing more advanced fusion strategy for better combining the features of clinical images and dermoscopy images, while ignore to explore the label relationship for further improvement.
Fu et al. \cite{Fu2022} introduced a graph relational model to utilize the relationship between diagnosis and SPC predictions. 
However, they did not integrate the prior information into the pipeline by introducing the graph model, so there is still room to investigate for further improvement.

\begin{figure*}[!t]
	\centering
	\includegraphics[width=16cm,height=7cm]{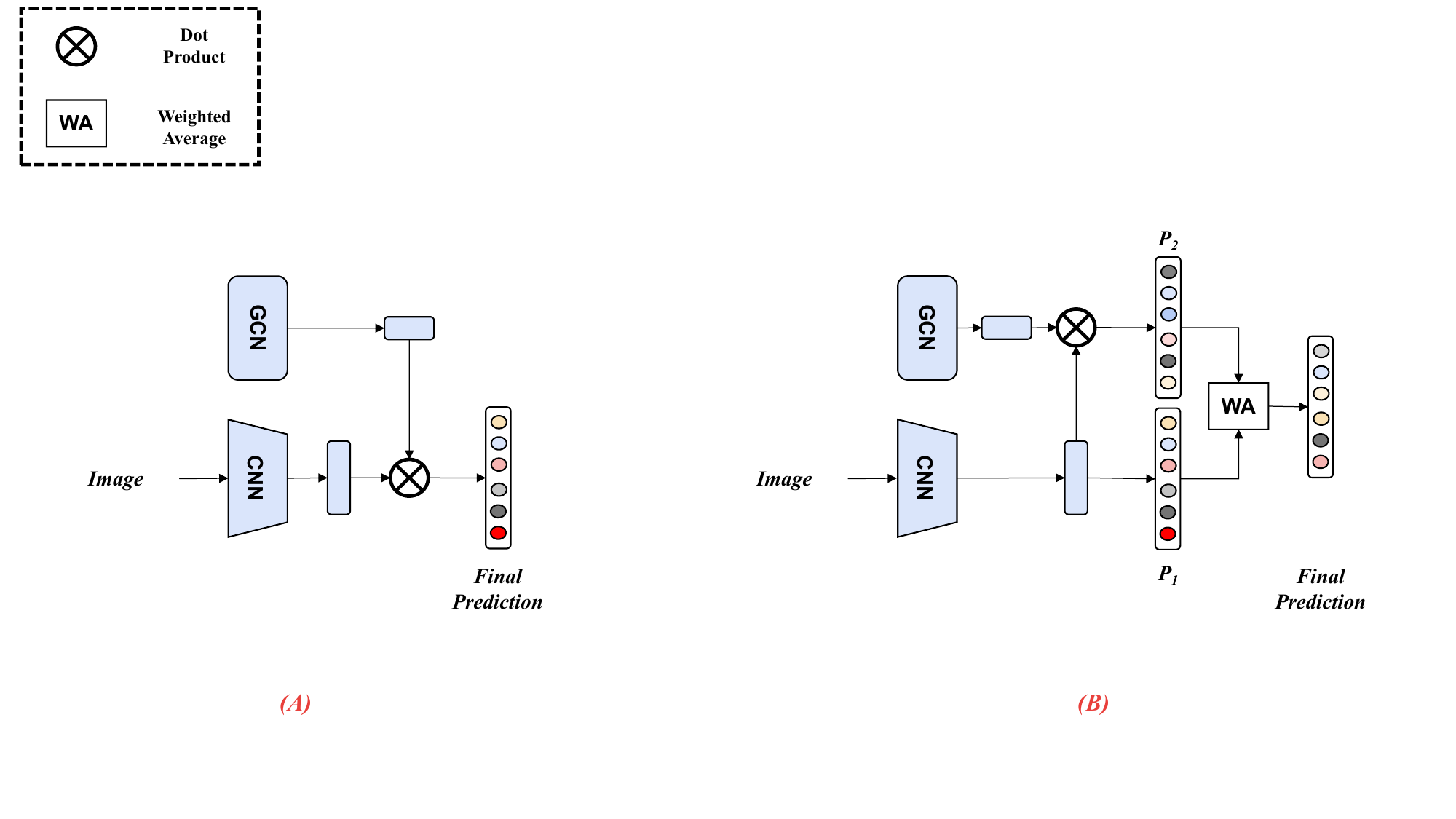}
	\vspace{-0.2in}
	\caption{The overview of (A) The model directly applying GCN and (B) our Graph-Ensemble Learning Network (GELN). GCN: Graph Convolutional Network, CNN: Convolutional Neural Network.}
	\label{fig_1}
\end{figure*}

Therefore, in this paper, we build a correlation matrix (CM) based on the co-occurrence times of each category and then integrate the prior information by introducing a graph convolutional network (GCN) and a label embedding module.
However, the GCN simply combines the features vector extracted from CNN and GCN for final prediction (See Fig.\ref{fig_1} (A)) and thus is easy be affected by the correlation matrix, especially in the case of rare medical data that makes statistical co-occurrences times may not match up with the correlations between some categories in the testing scenarios.
Collecting more samples is time- and labor-expensive, especially for the multi-modal medical data, and the prior information on label relationship will be helpful in enhancing multi-label classification accuracy.
Therefore, we aim to develop a new model, i.e., Graph-Ensemble Learning Network (GELN), to tackle this issue.
Unlike the commonly-used GCN (See Fig.\ref{fig_1} (A)) that takes the predictions using GCN as the final prediction, our GELN (see Fig.\ref{fig_1} (B)) considers it ($P_2$) as the complementary information of the prediction only from CNN ($P_1$) and then integrates them via a weighted averaging scheme.
In this ensemble learning way, our GELN is more robust in the different scenarios as it can selectively use the information based on label relationships.  

Our contributions can be summarized as follows:
1. our GELN integrates prior information of co-occurrence times into the multi-label skin lesion classification pipeline, which is different from previous methods.
2. our GELN adopts the idea of ensemble learning to solve the problem of GCN that is not well-generalized in the scenario of few medical data.
3. our GELN achieves state-of-the-art performance on the SPC dataset and consistently increases the accuracy of the fusion model on different datasets.

%
%
%
\subsection{Related Works}
Traditional automatic skin cancer classification methods \cite{Abbas2013,XieF2016,Celebi2008, Rajpara2009, Fabbrocini2014} mainly paid attention to extracting the representative hand-crafted features, e.g., shape and color, and then feed them into the existing machine learning classifiers, e.g., artificial neural network and support vector machine.
For instance, Xie et al. \cite{XieF2016} extracted the border, texture, and color features and then input them into the neural network model for classifying melanoma.
Fabbrocini et al. \cite{Fabbrocini2014} designed a processing pipeline based on a seven-point checklist to capture the representative features.
However, these methods suffer the generalization problem, especially for difficult cases, e.g., the case covered by hairs, as extracting the wanted hand-crafted features in these cases is challenging.

Recently, CNN-based methods \cite{Yu2017, Zhang2019, Tang2020, Liu2022} have been proposed in the classification challenge of International Skin Image Collaboration (ISIC) \cite{Codella2018,Celebi2019}. Yu et al. \cite{Yu2017} won first place in the ISIC 2016 classification challenge by designing a very deep residual network. Liu et al. \cite{Liu2022} proposed a clinical-inspired network by integrating the prior clinical practice of Zoom, observing and comparing steps into the deep learning pipeline.
However, these methods are still limited by only using single-modality data and are not trivial in exploring the use of multi-modal data \cite{Fu2022,Wang2022}.

More recently, many multi-modal methods \cite{Kawahara2019, Yap2018, Ge2017, Bi2020, Wang2022, Tang2022, Fu2022} that use clinical and dermoscopy images have been presented in the literature.
Kawahara et al. \cite{Kawahara2019} released the first multi-modal and multi-label skin lesion classification dataset, which laid a foundation for later multi-modal research.
Yap et al. \cite{Yap2018} presented an EmbeddingNet that first extracted the features from clinical and dermoscopy images and then fused them with metadata for classification.
Bi et al. \cite{Bi2020} introduced a HcCNN, which builds a hyper-branch module to integrate the multi-scale features from clinical and dermoscopy modalities.
Tang et al. \cite{Tang2022} proposed a FusionM4Net that did not fuse the feature vectors from these two modalities but conducted a late fusion scheme to integrate predictions from clinical, dermoscopy, and fusion branches.
Wang et al. \cite{Wang2022} designed an adversarial multi-modal scheme that introduces adversarial learning to better fuse the information from both modalities.
These methods focused on developing a more advanced module or scheme for combining multi-modal images. However, they did not consider exploiting the dependency relationship between the labels.
Fu et al. \cite{Fu2022} constructed a graph relational model to capture the relationship between categories and fuse the information of clinical and dermoscopy images.
However, this method seems more like a stacking method \cite{Wang2022, Cui2021, Cao2019} that only takes the predictions from images as input for learning more refined predictions than a graph method.
So it heavily relies on the accuracy of the first predictions from images and does not take advantage of the prior information of the categories, such as co-occurrence times, in the graph model.

Graph convolutional network (GCN) \cite{Wu2021, Kipf2017} has proven their effectiveness in the task of multi-label classification \cite{Wu2021, Kipf2017, Chen2019, Li2020,Meng2019, You2020,Wu2020S}.
GCN was first proposed on graph-structured data for semi-supervised classification \cite{Kipf2017}, in which objects/labels are viewed as nodes, and the relationships between nodes are considered as edges. Then convolutions are used to extract the features between adjacent nodes.
Chen et al. \cite{Chen2019} constructed a GCN by modeling the prior label dependencies (co-occurrence times) and integrating it with the image feature vector extracted from CNN.
Other methods \cite{Li2020, Meng2019, You2020, Wu2020S} followed the structures with GCN \cite{Chen2019} and added more advanced modules into the pipeline, such as the attention module \cite{Meng2019}.
However, these methods are vulnerable to the inaccurate correlation matrix and thus are not suitable to be applied in the case of multi-modal medical data, whose statistical samples are rare.
\begin{figure*}[t]
	\centering
	\includegraphics[width=18cm,height=9cm]{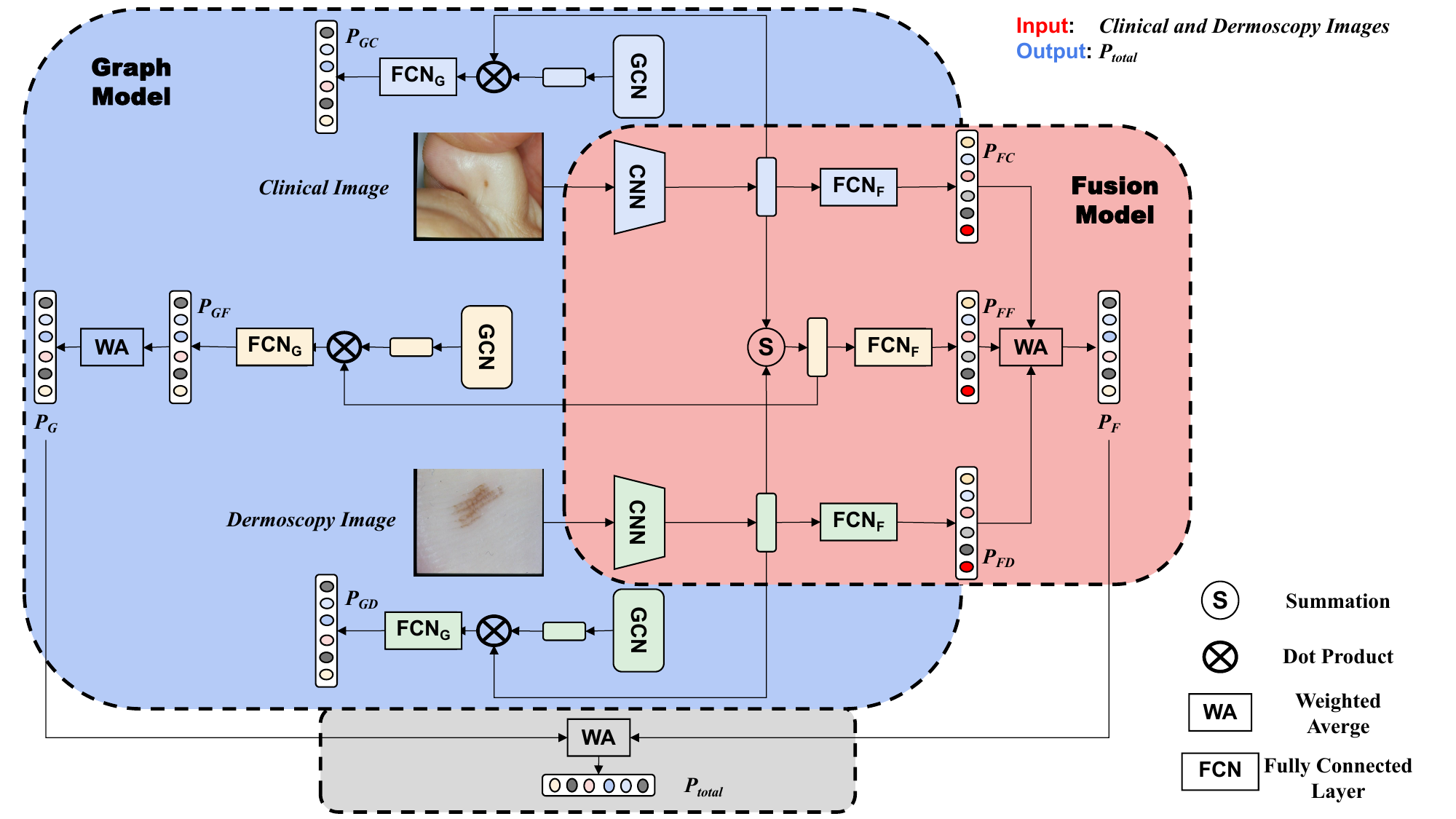}
	\caption{The pipeline of our Graph-Ensemble Learning Network.
	GCN: Graph Convolutional Network, CNN: Convolutional Neural Network, $FCN_F$: Fully connected layers in the fusion model, $FCN_G$: Fully connected layers in the graph model.}
	\label{fig_2}
\end{figure*}
\section{Our method: Graph Ensemble Learning Network (GELN)}
As shown in Fig.\ref{fig_2},  our GELN mainly consists of three parts, i.e., a multi-modal fusion model, a graph model and the weighted averaging scheme.
Firstly, we train a fusion model to obtain the prediction only based on the features extracted from clinical and dermoscopy images.
Then, we embed the GCN into the fusion model that combines the features prior correlation of categories and the image features extracted from fusion model to get the another prediction. 
Finally, we utilize a weighted averaging scheme to fuse above-mentioned two predictions, which can utilizes the useful information from GCN while avoiding its negative influences as much as possible.
In the following text, we will introduce the fusion model, graph convolutional model and the details of training and testing of our method in sequence.

\subsection{Fusion Model}
In our GELN, we apply the FusionM4Net \cite{Tang2022} as our multi-modal fusion model (the red part in Fig.\ref{fig_2}), as it achieved excellent performance and released the code and other previous methods with SOTA performance do not.
The fusion model uses two ResNet-50 \cite{He2016} as image feature extractors that capture the visual features from clinical and dermoscopy images. 
The features fr
A fused feature vector is obtained using a summation operation to fuse extracted clinical and dermoscopy features. 
Then, three fully connected layers $ FCN_F$ are used to predict these three features.
Finally, three predictions, i.e., $P_{FC}$, $P_{FD}$ and $P_{FF}$, are weighted averaged to formulate the prediction of $P_F$.
The fusion model provides the prediction $P_F$ only from CNN, th, and pre-trained weights of CNNs to support the following operations of our GELN.

\subsection{Graph Model}
From Fig.\ref{fig_2}, we can see that the graph model contains three GCNs, which combine the features from different branches, i.e., clinical image branch, dermoscopy image branch, and fused feature branch. 
Then, each GCN is followed with a $FCN_G$ to obtain the prediction.
$FCN_G$ is comprised of a weighted-shared part, i.e., two fully connected (FC) layers, Batch normalization (BN) \cite{loffe2015} layers and Swish activation layers \cite{Ramachandran2017}, to learn the representation of the categories features, and contain multiple classifiers $FC_i$ ($i \in {1,2,...,N}, N=8$) to obtain the predictions of multi-categories $P_i \in \mathbb{R}^{1 \times K_i}$  ($i \in {1,2,...,N}, N=8$), where $K_i$ means the number of classes in the category $i$. More details about the categories can be seen in Table \ref{table_1}.
Finally, a weighted average scheme is adopted to selectively fuse these predictions from the GCNs, which naturally deals with the problem that commonly-used GCNs are not suitable in the multimodal scenarios, as they do not think about how to utilize the modality-unique characteristics when exploiting the inter-categorical relationship, or how much influence of each modality should put on the final prediction \cite{Fu2022}. 

\begin{figure}[t]
	\centering
	\includegraphics[width=9cm,height=5cm]{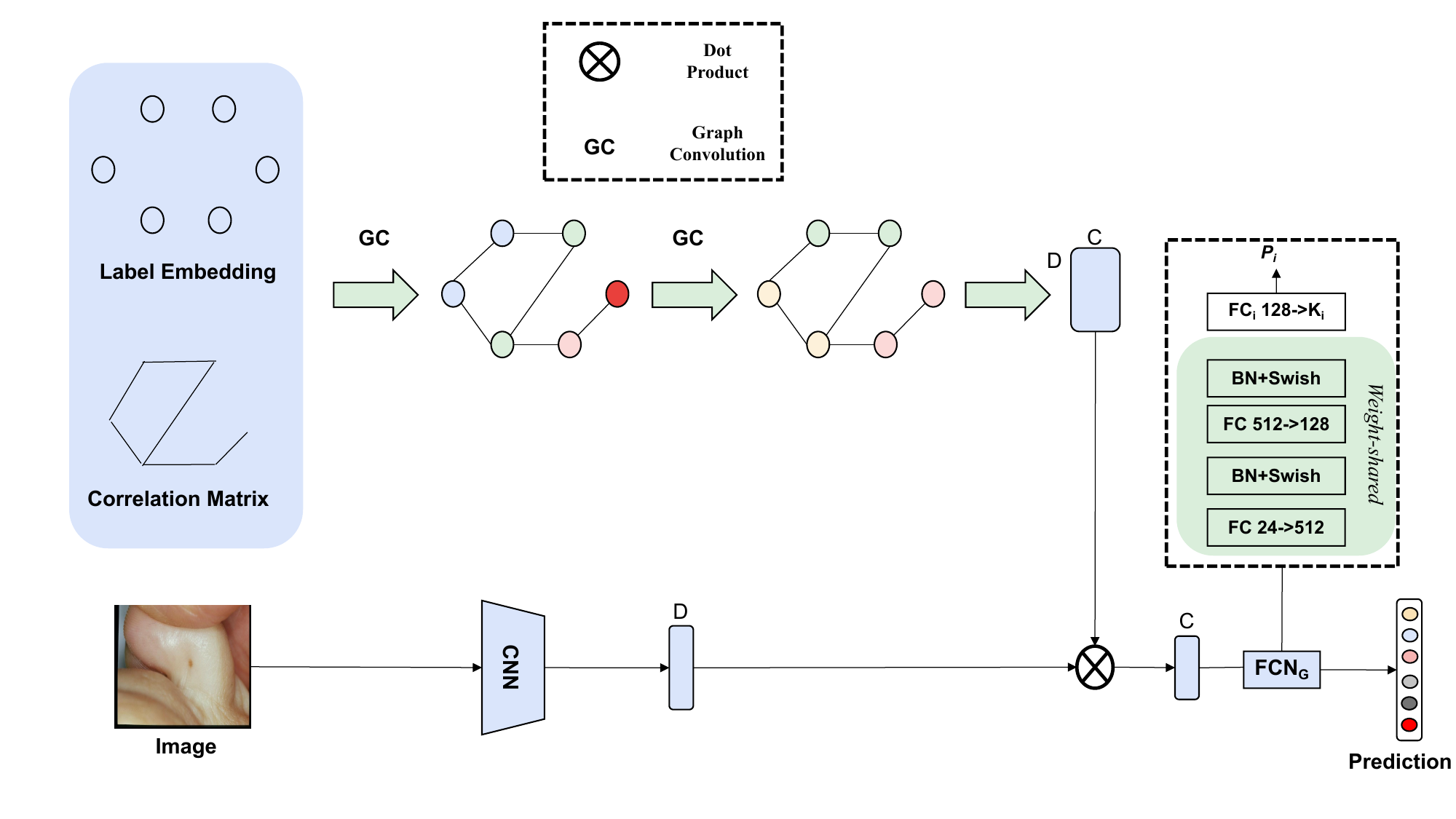}
	\caption{How does GCN work in the proposed GELN.}
	\label{fig_3}
\end{figure}

\subsubsection{Graph Convolution Network}
The main component of the graph model is Graph Convolution Network (GCN). Therefore, we recap how GCN work.
In total, a graph convolutional layer can be represented as Eq.\ref{eq1}:
\begin{equation}\label{eq1}
	\boldsymbol{F}^{j+1}=f\left(\boldsymbol{CM} \boldsymbol{F}^j \boldsymbol{W}^l\right)
\end{equation}
where $\boldsymbol{CM} \in \mathbb{R}^{n \times n}$ denotes the correlation matrix, $\boldsymbol{F}^j \in \mathbb{R}^{n \times d}$ indicates the extracted features, and $\boldsymbol{W}^l \in \mathbb{R}^{d \times d_1}$ is a learned transformation function (where $n$ denotes the number of node and the $d$ indicates the dimension number of node features). $f(\cdot)$ is a non-linear activation function.
The object of GCN is to update the $\boldsymbol{F}^{j+1} \in \mathbb{R}{C \times D}$ by taking $\boldsymbol{CM} \in \mathbb{R}^{n \times n}$ and  $\boldsymbol{F}^j \in \mathbb{R}^{n \times d}$ as input.

More concretely, in the task of multi-label skin lesion classification, we use two-stacked graph convolutional layers to construct the GCN, where the input of each layer of GCN is the node features from the last layer $\boldsymbol{F}^j$ and output the updated node features $\boldsymbol{F}^{j+1}$.
In the first layer, it takes the label features $\boldsymbol{LF} \in \mathbb{R}^{C \times d}$ obtained by word embedding, where $C$ is the number of the whole class in the multi-label task and d is the dimension of each label-embedding feature.
In the last layer, it outputs the learned node feature representation $Z \in \mathbb{R}^{C \times D}$, where D is the dimension of the extracted image features representation.

\subsubsection{Correlation Matrix in GCN}
\begin{figure*}[t]
	\centering
	\includegraphics[width=9cm,height=10cm]{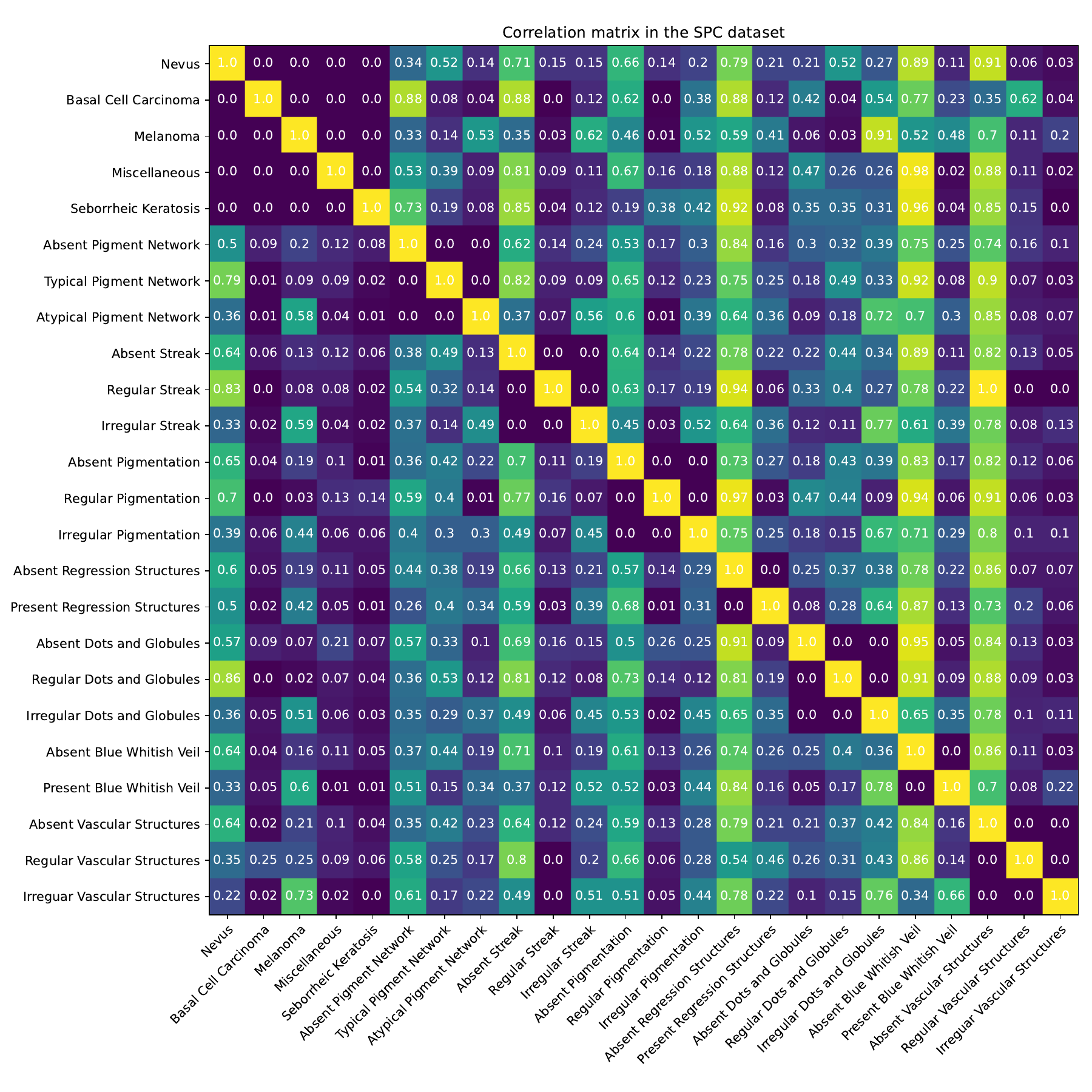}
	\includegraphics[width=9cm,height=10cm]{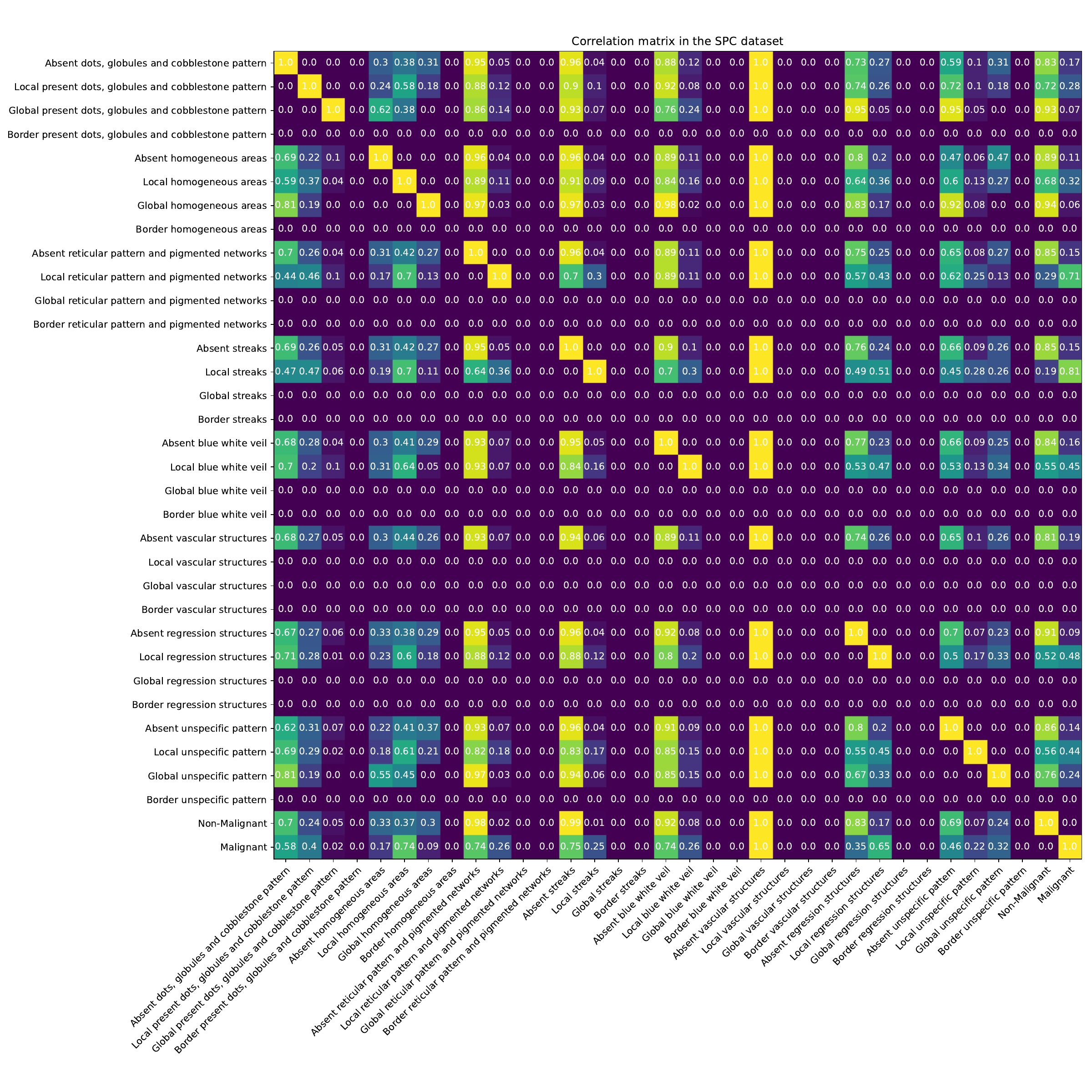}
	
	\caption{The correlation matrix in the SPC dataset (left) and the ISIC-2016 dataset (right).}
	\label{fig_4}
\end{figure*}

The core component of GCN is the correlation matrix (CM) (or called the adjacent matrix), which is used as the edges to propagate the information between nodes.
CM is predefined in most tasks while given in the multi-label classification task. 
Therefore, we follow the former GCN methods \cite{Chen2019, Li2020, Meng2019} that build the CM by exploiting conditional probability $p(L_j|L_i)$, which indicates the probability of label $L_j$ appearing when label $C_i$ appears.
Based on the definition of $p(C_j|C_i)$, we formulate the CM based on the co-occurrence of each class in the dataset.
Firstly, the occurrences of all the label pairs in the training set (including training and validation set), and we define $M_{ij}$ as the co-occurrence times of $L_i$ and $L_j$ and $M_i$ as the occurrence time of $L_i$. So, we can get the conditional probability based on Eq.\ref{eq2}:
\begin{equation}\label{eq2}
	p(L_j|L_i) = \frac{p(L_j,L_i)}{p(L_i)} = \frac{M_{ij}}{M_i}
\end{equation}
where $p(L_j,L_i)$ means the probability of co-occurrence of $L_i$ and $L_j$ and $p(L_i)$ denotes the probability of occurrence of $L_i$.
So, the label pairs' conditional probability can be computed based on Eq.\ref{eq2}, and thus, a conditional probability can be built.
The correlation matrix of our experiments can be seen in Fig.~\ref{fig_4}.
\subsection{Others}
There are steps in the implementation of our GELN.
In the first step, we only update the weight of the fusion model $\theta_{F}$ by minimizing the loss $L^F$.


\begin{equation}\label{eq4}
	L_F = L_{FC} + L_{FD} + L_{FF}
\end{equation}
where the $L_{FC}$, $L_{FD}$ and $L_{FF}$ are the losses from the clinical image branch $P_{FC}$, dermoscopy image branch $P_{FD}$ and fused feature branch $P_{FF}$, respectively.

In the second step, we optimize our GELN by minimizing $L_G$:
\begin{equation}\label{eq5}
	L_G = L_{GC} + L_{GD} + L_{GF}
\end{equation}
where the $L_{GC}$, $L_{GD}$ and $L_{GF}$ are the losses from the graph branches of $P_{GC}$, $P_{GD}$ and $P_{GF}$, respectively.
During the training of the second step, we can obtain two versions of our GELN, i.e., GELN-freeze and GELN-unfreeze,  by two settings.
The first one is initializing the fusion model with the pre-trained weight in the first step and freezing the fusion model.
The second one is not initializing the fusion model with the pre-trained weight in the first step and freezing the fusion model.
In the last step, we search for the weight with the best performance in the validation dataset and then fuse the $P_F$ and $P_G$ using the searched weights. We refer \cite{Tang2022} to know more details about the weighted average scheme.

\begin{table*}[t]
	\centering
	\caption{Performance comparison of our GELN-freezed and GELN-unfreezed with previous methods in the SPC dataset, in terms of AUC ($\%$). Avg: Average}
	\begin{tabular}{ccccccccc}
		\hline
		\multirow{2}{*}{Category} & \multirow{2}{*}{class} & \multirow{2}{*}{GELN-freezed}  & \multirow{2}{*}{GELN-unfreezed} & \multirow{2}{*}{GIIN}          & \multirow{2}{*}{AMFAM}         & \multirow{2}{*}{FM-FS}         & \multirow{2}{*}{HcCNN} & \multirow{2}{*}{Inception-com} \\
		&                        &                                &                                 &                                &                                &                                &                        &                                \\ \hline
		\multirow{10}{*}{Diag}    & \multirow{2}{*}{BCC}   & \multirow{2}{*}{\textbf{95.8}} & \multirow{2}{*}{95.5}           & \multirow{2}{*}{92.8}          & \multirow{2}{*}{94.1}          & \multirow{2}{*}{95.3}          & \multirow{2}{*}{94.4}  & \multirow{2}{*}{92.9}          \\
		&                        &                                &                                 &                                &                                &                                &                        &                                \\
		& \multirow{2}{*}{NEV}   & \multirow{2}{*}{92.4}          & \multirow{2}{*}{\textbf{92.7}}  & \multirow{2}{*}{86.8}          & \multirow{2}{*}{89.7}          & \multirow{2}{*}{92.6}          & \multirow{2}{*}{87.7}  & \multirow{2}{*}{89.7}          \\
		&                        &                                &                                 &                                &                                &                                &                        &                                \\
		& \multirow{2}{*}{MEL}   & \multirow{2}{*}{90}            & \multirow{2}{*}{\textbf{90.2}}  & \multirow{2}{*}{87.6}          & \multirow{2}{*}{89.1}          & \multirow{2}{*}{89}            & \multirow{2}{*}{85.6}  & \multirow{2}{*}{86.3}          \\
		&                        &                                &                                 &                                &                                &                                &                        &                                \\
		& \multirow{2}{*}{MISC}  & \multirow{2}{*}{93.2}          & \multirow{2}{*}{92.4}           & \multirow{2}{*}{88.8}          & \multirow{2}{*}{90.6}          & \multirow{2}{*}{\textbf{94.1}} & \multirow{2}{*}{88.3}  & \multirow{2}{*}{88.3}          \\
		&                        &                                &                                 &                                &                                &                                &                        &                                \\
		& \multirow{2}{*}{SK}    & \multirow{2}{*}{88.9}          & \multirow{2}{*}{90.0}           & \multirow{2}{*}{79.8}          & \multirow{2}{*}{81.7}          & \multirow{2}{*}{89.2}          & \multirow{2}{*}{80.4}  & \multirow{2}{*}{\textbf{91.0}} \\
		&                        &                                &                                 &                                &                                &                                &                        &                                \\
		\multirow{4}{*}{PN}       & \multirow{2}{*}{TYP}   & \multirow{2}{*}{83.2}          & \multirow{2}{*}{84.0}           & \multirow{2}{*}{80.1}          & \multirow{2}{*}{84.5}          & \multirow{2}{*}{85.9}          & \multirow{2}{*}{85.9}  & \multirow{2}{*}{\textbf{84.2}} \\
		&                        &                                &                                 &                                &                                &                                &                        &                                \\
		& \multirow{2}{*}{ATP}   & \multirow{2}{*}{85.1}          & \multirow{2}{*}{\textbf{86.3}}  & \multirow{2}{*}{87.5}          & \multirow{2}{*}{82}            & \multirow{2}{*}{83.9}          & \multirow{2}{*}{78.33} & \multirow{2}{*}{79.9}          \\
		&                        &                                &                                 &                                &                                &                                &                        &                                \\
		\multirow{4}{*}{STR}      & \multirow{2}{*}{REG}   & \multirow{2}{*}{89.2}          & \multirow{2}{*}{\textbf{90.4}}  & \multirow{2}{*}{84.9}          & \multirow{2}{*}{89.5}          & \multirow{2}{*}{87.9}          & \multirow{2}{*}{87.8}  & \multirow{2}{*}{87}            \\
		&                        &                                &                                 &                                &                                &                                &                        &                                \\
		& \multirow{2}{*}{IR}    & \multirow{2}{*}{81.9}          & \multirow{2}{*}{\textbf{82.3}}  & \multirow{2}{*}{81.2}          & \multirow{2}{*}{80.7}          & \multirow{2}{*}{81.4}          & \multirow{2}{*}{77.6}  & \multirow{2}{*}{78.9}          \\
		&                        &                                &                                 &                                &                                &                                &                        &                                \\
		\multirow{4}{*}{PIG}      & \multirow{2}{*}{REG}   & \multirow{2}{*}{82.2}          & \multirow{2}{*}{\textbf{85.3}}  & \multirow{2}{*}{81.1}          & \multirow{2}{*}{85.1}          & \multirow{2}{*}{80.9}          & \multirow{2}{*}{83.6}  & \multirow{2}{*}{74.9}          \\
		&                        &                                &                                 &                                &                                &                                &                        &                                \\
		& \multirow{2}{*}{IR}    & \multirow{2}{*}{83.5}          & \multirow{2}{*}{\textbf{83.8}}  & \multirow{2}{*}{83.6}          & \multirow{2}{*}{83.4}          & \multirow{2}{*}{83.5}          & \multirow{2}{*}{81.3}  & \multirow{2}{*}{79}            \\
		&                        &                                &                                 &                                &                                &                                &                        &                                \\
		\multirow{2}{*}{RS}       & \multirow{2}{*}{PRS}   & \multirow{2}{*}{82.0}          & \multirow{2}{*}{82.9}           & \multirow{2}{*}{79}            & \multirow{2}{*}{\textbf{86.7}} & \multirow{2}{*}{81.7}          & \multirow{2}{*}{81.9}  & \multirow{2}{*}{82.9}          \\
		&                        &                                &                                 &                                &                                &                                &                        &                                \\
		\multirow{4}{*}{DaG}      & \multirow{2}{*}{REG}   & \multirow{2}{*}{78.8}          & \multirow{2}{*}{\textbf{80.6}}  & \multirow{2}{*}{78.6}          & \multirow{2}{*}{77.7}          & \multirow{2}{*}{79.1}          & \multirow{2}{*}{77.7}  & \multirow{2}{*}{76.5}          \\
		&                        &                                &                                 &                                &                                &                                &                        &                                \\
		& \multirow{2}{*}{IR}    & \multirow{2}{*}{81.2}          & \multirow{2}{*}{82.0}           & \multirow{2}{*}{\textbf{83.1}} & \multirow{2}{*}{81.9}          & \multirow{2}{*}{80.1}          & \multirow{2}{*}{82.6}  & \multirow{2}{*}{79.9}          \\
		&                        &                                &                                 &                                &                                &                                &                        &                                \\
		\multirow{2}{*}{BWV}      & \multirow{2}{*}{PRS}   & \multirow{2}{*}{91.5}          & \multirow{2}{*}{\textbf{91.8}}  & \multirow{2}{*}{90.8}          & \multirow{2}{*}{91.1}          & \multirow{2}{*}{90.6}          & \multirow{2}{*}{89.8}  & \multirow{2}{*}{89.2}          \\
		&                        &                                &                                 &                                &                                &                                &                        &                                \\
		\multirow{4}{*}{VS}       & \multirow{2}{*}{REG}   & \multirow{2}{*}{88.6}          & \multirow{2}{*}{88.5}           & \multirow{2}{*}{80.7}          & \multirow{2}{*}{\textbf{88.8}} & \multirow{2}{*}{87.8}          & \multirow{2}{*}{87.0}  & \multirow{2}{*}{85.5}          \\
		&                        &                                &                                 &                                &                                &                                &                        &                                \\
		& \multirow{2}{*}{IR}    & \multirow{2}{*}{84.0}          & \multirow{2}{*}{\textbf{84.5}}  & \multirow{2}{*}{75.4}          & \multirow{2}{*}{80.9}          & \multirow{2}{*}{78.9}          & \multirow{2}{*}{82.7}  & \multirow{2}{*}{76.1}          \\
		&                        &                                &                                 &                                &                                &                                &                        &                                \\
		\multirow{2}{*}{Avg}      & \multirow{2}{*}{}      & \multirow{2}{*}{86.5}          & \multirow{2}{*}{\textbf{87.1}}  & \multirow{2}{*}{83.6}          & \multirow{2}{*}{85.7}          & \multirow{2}{*}{86}            & \multirow{2}{*}{84.3}  & \multirow{2}{*}{83.7}          \\
		&                        &                                &                                 &                                &                                &                                &                        &                                \\ \hline
	\end{tabular}
	\label{table_2}
	
\end{table*}

\section{Experiments}
\subsection{Dateset and metrics}
We used two public datasets to evaluate our method, including the seven-point checklist (SPC) dataset \cite{Kawahara2019} and ISIC 2016 dataset \cite{David2016}.
The SPC dataset is well-recognized \cite{Fu2022,Tang2022,Bi2020,Kawahara2019} in the multi-label skin lesion classification task. It totally contains 1033 cases and is divided into 413 training, 203 validation, and 395 testing cases.
Each case includes clinical and dermoscopy images and multi-categories ground truth, and each category has different classes (See Table~\ref{table_1}).

The ISIC 2016 dataset contains 900 training cases and 379 testing cases.
Each case only includes the dermoscopy image and the multi-categories ground truth, i.e., Dots, globules and Cobblestone pattern, Homogeneous areas, Reticular pattern and pigmented networks, Streaks, Blue-white veil, Vascular structures, Regression structures, Unspecific pattern, Diagnosis.
Diagnosis only includes two classes: melanoma and non-melanoma, and the other eight categories all consist of 4 classes: Absent, Local present, Border present, and Global present.

We followed the previous multi-label methods \cite{Kawahara2019,Bi2020,Fu2022} and used the area under the receiver operating curve (AUC, main ranking metric), precision (Prec), Specificity (Spc), and Sensitivity (Sen) for the evaluation and comparison in our experiments.
\begin{table}[t]
		\centering
		\caption{The class details of each category in the Seven Point Checklist dataset}
			\scalebox{0.8}{
		\renewcommand\arraystretch{1.75}
		\setlength{\tabcolsep}{1mm}{
	\begin{tabular}{c|cccccccccccc}
		\hline
		\textit{\textbf{Category}} & \multicolumn{5}{c|}{\textbf{Diagnosis}}                              & \multicolumn{3}{c|}{\textbf{PN}}                         & \multicolumn{2}{c|}{\textbf{BWV}}                  & \multicolumn{2}{c}{\textbf{RS}} \\ \hline
		\textit{\textbf{Class}}    & BCC & NEV & MEL                     & MISC & \multicolumn{1}{c|}{SK} & ABS                     & TYP & \multicolumn{1}{c|}{ATP} & ABS                     & \multicolumn{1}{c|}{PRS} & ABS            & PRS            \\ \hline
		\textit{\textbf{Category}} & \multicolumn{3}{c|}{\textbf{VS}}    & \multicolumn{3}{c|}{\textbf{PIG}}                        & \multicolumn{3}{c|}{\textbf{STR}}                        & \multicolumn{3}{c}{\textbf{DaG}}                           \\ \hline
		\textit{\textbf{Class}}    & ABS & REG & \multicolumn{1}{c|}{IR} & ABS  & REG                     & \multicolumn{1}{c|}{IR} & ABS & REG                      & \multicolumn{1}{c|}{IR} & ABS                      & REG            & IR             \\ \hline
	\end{tabular}}}
\label{table_1}
\end{table}

\subsection{Implementation details}
We train the GELN using 250 epochs using an Adam optimizer \cite{Kingma2014} with a batch size of 32 and a learning rate of 3e-5.
The learning rate is reduced by a CosineAnnealing schedule.
During our training, the stochastic weights averaging scheme (SWA) \cite{Pavel2018} is used in the last 50 epochs.
During the training of the fusion model, we initialize the ResNet-50 with the pre-trained weight on ImageNet \cite{Deng2009}.
Data augmentations, including horizontal and vertical flipping, shift, rotation, brightness, and contrast enhancement, are randomly employed during the training.
The hyperparameter setting of all the training and testing of our method are the same unless specified and under the configuration of a GPU 100 with 40GB.
Our model and its variant are built based on Pytorch.
We train our GELN 50 times, i.e., fusion model and graph model 50 times, respectively, to get the average value and standard deviation for our ablation study.
While in comparison with the previous method, we display the model with the highest AUC value, as the previous method only shows the experimental result of a single time, and we suppose their results are with the best performance in their experiments.

\subsection{Comparison with previous methods}
We compared with our method with previous state-of-the-art methods, i.e., Inception-combined methods \cite{Kawahara2019}, EmbeddingNet \cite{Yap2018}, TripleNet \cite{Ge2017}, HcCNN \cite{Bi2020}, AMFAM \cite{Wang2022}, FusionM4Net-FS \cite{Tang2022} and GIIN \cite{Fu2022} on the SPC dataset.
The results of these methods are quoted from their papers or the results from \cite{Bi2020, Fu2022}.
We first conduct the performance comparison between our methods and existing state-of-the-art methods on the SPC dataset, based on the AUC value of all the categories, and display the results in Table~\ref{table_2}. Then, we show the performance comparison of detecting visual characteristics of Melanoma between all the methods in Table~\ref{table_3}

From Table~\ref{table_2}, we can see that the two variants of our GELN, i.e., GELN-freeze (86.5$\%$) and GELN-unfreeze (87.1$\%$), outperform other methods based on the averaged AUC value.
Our GELN achieves the highest averaged AUC value of 87.1$\%$ and consistently performs better than previous methods in 10 out of 17 classes. Also, it reaches the highest AUC in the class of Melanoma and the highest averaged AUC value of 92.16$\%$ in the category of diagnosis, which is slightly better than GELN-freeze (92.06$\%$) and FM-FS (92.04$\%$).
In the comparison of previous methods without using graph model, i.e., AMFAM \cite{Wang2022}, FusionM4Net-FS (FM-FS) \cite{Tang2022}, HcCNN \cite{Bi2020} and Inception-combined \cite{Kawahara2019}, FM-FS (86.0$\%$)and AMFAM (85.7$\%$) significantly outperforms the HcCNN ($84.3\%$) and Inception-combined (83.7$\%$) as they not only focused on developing a more effective fusion module like HcCNN and Inception-combined but also utilize the learning schemes, such as adversarial training in AMFAM and ensemble learning scheme in FM-FS, to increase the classification accuracy.
In the comparison between the methods using the graph model, i.e., GELN methods and GIIN, both our GELN-freezed and GELN-unfreezed, outperforms GIIN by $2.8\%$ and $3.4\%$, respectively.
We attribute this to the advantage of our method that efficiently integrates the prior co-occurrence information in our classification pipeline.
Furthermore, we are surprised that GIIN achieves the worst performance of 83.6$\%$ in Table~\ref{table_2}; we think it is because although efficiently utilizing the relationship of inter-categories is crucial in enhancing the accuracy of multi-label tasks, while its effect is not as good as the other models, such as AMFAM, HcCNN, and FM-FS, in solving the problem of combining the clinical and dermoscopy image for multi-label classification.  
These results demonstrate that taking advantage of the label dependencies (relationship) can improve the multi-label classification performance, but how much it can improve depends on how to use it.


From Table~\ref{table_3}, we can observe that the proposed GELN-unfreeze obtains the highest value both in the metrics of averaged AUC (85.2$\%$) and averaged sensitivity (65.3$\%$) across the eight categories i.e., Melanoma and other seven-point checklist categories contributes to the score for the diagnosis of Melanoma.
It also gets the best AUC (90.2$\%$) value and second-best Sen value (68.1 $\%$) in the category of Melanoma,  which illustrates the advantage of our method in detecting Melanoma, the most dangerous type in the skin lesion.
Then, we observe some interesting phenomena in these comparison results; that is, almost all the models achieve a high Spe value and low Sen value. For instance, in the category of irregular vascular structure (VS-IR), the Sen values of all the methods are not greater than 50$\%$, but the corresponding Spe value is all greater than 92.6$\%$, especially for GIIN, whose Spe value is 100$\%$.
These results are extremely biased to the negative class, as the unbalanced distribution of the SPC dataset \cite{Kawahara2019}, e.g., there are only a few VS-IR (71 cases) among the total 1011 cases, makes all the models tend to overfit the dominant cases.
Also, we can find that the Sen and Spe values of BWV-PRS and RS-PRS of our methods are more balanced than PIG-IR and STR-IR, even  BWV-PRS and RS-PRS are also with a more unbalanced classes distribution, i.e., the number of positive PRS case for BWV is 195 (negative is 816) and the number of positive PRS for RS is 253 (negative is 758), than PIG-IR and STR-IR, i.e., the number of positive IR case for PIG is 305 (negative is 706 ) and the number of positive IR for STR is 251 (negative is 760).
It is because the category containing two classes (BWV and RS) is easier to classify than the category containing three classes (PIG and STR).

\begin{table*}[t]
	\centering
	\caption{The performance comparison of detecting visual charactersitics of melanoma ($\%$). Avg: Average}
	\begin{tabular}{ccccccccccc}
		\hline
		\multirow{2}{*}{}     & \multirow{2}{*}{Method} & DIAG          & PN            & STR           & PIG           & RS            & DaG           & BWV           & VS            & \multirow{2}{*}{Avg} \\
		&                         & MEL           & ATP           & IR            & IR            & PRS           & IR            & PRS           & IR            &                      \\ \hline
		\multirow{9}{*}{AUC}  & Inception-combined      & 86.3          & 79.9          & 78.9          & 79            & 82.9          & 79.9          & 89.2          & 76.1          & 81.5                 \\
		& EmbeddingNet            & 82.5          & 74.5          & 77.7          & 77.9          & 71.3          & 78.5          & 84.8          & 76.9          & 78                   \\
		& TripleNet               & 81.2          & 73.8          & 76.3          & 77.6          & 76.8          & 76            & 85.1          & 79.9          & 78.4                 \\
		& HcCNN                   & 85.6          & 78.3          & 77.6          & 81.3          & 81.9          & 82.6          & 89.8          & 82.7          & 82.5                 \\
		& FusionNet-FS            & 89            & 83.9          & 81.4          & 83.5          & 81.7          & 80.1          & 90.6          & 78.9          & 83.7                 \\
		& AMFAM                    & 89.1          & 82            & 80.7          & 83.4          & 86.7          & 81.9          & 91.1          & 80.9          & 84.4                 \\
		& GIIN                    & 87.6          & \textbf{87.5} & 81.2          & 83.6          & 79            & \textbf{83.1} & 90.8          & 75.4          & 83.5                 \\
		& GELN-freezed            & 90            & 83.2          & 81.9          & 83.2          & 82            & 81.2          & 91.5          & 84.0          & 84.7                 \\
		& GELN-unfreezed          & \textbf{90.2} & 84            & \textbf{82.3} & \textbf{83.8} & \textbf{82.9} & 81.9          & \textbf{91.8} & \textbf{84.5} & \textbf{85.2}        \\ \hline
		\multirow{9}{*}{Prec} & Inception-combined      & 65.3          & 61.6          & 52.7          & 57.8          & 56.5          & 70.5          & 63            & 30.8          & 57.3                 \\
		& EmbeddingNet            & 68.3          & 52.5          & \textbf{58.5} & 61.5          & 76.8          & 70.8          & 89.5          & 36.8          & 64.4                 \\
		& TripleNet               & 61.8          & 59.6          & 61.7          & 54.7          & 77.7          & 65.7          & 90.6          & 64.3          & 67                   \\
		& HcCNN                   & 62.8          & 62.3          & 52.4          & 65.1          & 81.6          & 69.6          & \textbf{91.9} & 50            & 67                   \\
		& FusionNet-FS            & 65.7          & \textbf{82.2} & 56.2          & 67.6          & \textbf{82}   & 67.2          & 64.9          & 42.9          & 68.5                 \\
		& AMFAM                    & \textbf{76.2} & 51            & 54.3          & 61.3          & 46.2          & 82.5          & 56            & 0             & 53.4                 \\
		& GIIN                    & 65.6          & 48.4          & 50.4          & \textbf{82.3} & 73.5          & \textbf{74.9} & 67.4          & \textbf{100}  & \textbf{70.3}        \\
		& GELN-freezed            & 64.4          & 49.5          & 51.1          & 54.8          & 40.6          & 70.1          & 70.7          & 7             & 51                   \\
		& GELN-unfreezed          & 63.4          & 41.9          & 47.9          & 57.3          & 41.5          & 72.3          & 61.3          & 3             & 48.6                 \\ \hline
		\multirow{9}{*}{Sen}  & Inception-combined      & 61.4          & 48.4          & 51.1          & 59.7          & 66            & 62.1          & 77.3          & 13.3          & 54.9                 \\
		& EmbeddingNet            & 40.6          & 33.3          & 51.1          & 60.5          & 96.2          & 64.4          & 96.3          & 23.3          & 58.2                 \\
		& TripleNet               & 46.5          & 33.3          & 39.4          & 61.3          & 97.9          & 67.2          & 90            & 30            & 58.2                 \\
		& HcCNN                   & 58.4          & 40.9          & 35.1          & 55.7          & 95.2          & 80.2          & 92.2          & 20            & 59.7                 \\
		& FusionNet-FS            & 62.4          & 49.5          & 47.9          & 58.9          & 47.1          & 68.4          & 66.7          & 20            & 52.6                 \\
		& AMFAM                    & 65.8          & 58.5          & 57.3          & 67.9          & 72.1          & 66.7          & 75            & 0             & 57.9                 \\
		& GIIN                    & 59            & \textbf{77.5} & \textbf{67}   & 39.2          & 21.9          & 70.1          & 69.9          & 3.6           & 51                   \\
		& GELN-freezed            & \textbf{69.9} & 60.5          & 58.5          & \textbf{69.4} & 71.7          & \textbf{72.1} & 68.8          & 40            & 63.9                 \\
		& GELN-unfreezed          & 68.1          & 57.4          & 57            & 68.9          & \textbf{74.6} & 70            & \textbf{76.7} & \textbf{50}   & \textbf{65.3}        \\ \hline
		\multirow{9}{*}{Spe}  & Inception-combined      & 88.8          & 90.7          & 85.7          & 80.1          & 81.3          & 78.9          & 89.4          & 97.5          & 86.6                 \\
		& EmbeddingNet            & \textbf{93.5} & 90.7          & 88.7          & 82.7          & 20.8          & 78.4          & 52            & 96.7          & 75.4                 \\
		& TripleNet               & 90.1          & 93            & \textbf{92.4} & 76.8          & 23.6          & 71.6          & 60            & 98.6          & 75.8                 \\
		& HcCNN                   & 88.1          & \textbf{92.4} & 90            & 86.3          & 41.5          & 71.6          & 65.3          & 98.4          & 79.2                 \\
		& FusionNet-FS            & 88.8          & 90.1          & 88.4          & 88.1          & 96.2          & 72.9          & 91.6          & 97.8          & \textbf{89.2}        \\
		& AMFAM                    & 91.4          & 85.6          & 85.9          & 83            & 82.6          & 82.4          & 90.3          & 92.4          & 86.7                 \\
		& GIIN                    & 89.5          & 79            & 80.3          & \textbf{95.8} & \textbf{96.8} & 78.8          & 91            & \textbf{100}  & 88.9                 \\
		& GELN-freezed            & 88.1          & 82.3          & 81.1          & 81.2          & 76.2          & \textbf{93.1} & \textbf{92.8} & 92.8          & 85.4                 \\
		& GELN-unfreezed          & 87.7          & 83.5          & 84.5          & 81.9          & 81.6          & 76.8          & 91.3          & 92.6          & 85.0                 \\ \hline
	\end{tabular}
	\label{table_3}
\end{table*}

\begin{table}[h]
	\centering
	\caption{Ablation study of our GELN methods, including GELN-freezed and GELN-unfreezed in terms of AUC ($\%$).}

	\begin{tabular}{cccc}
		\hline
		\multicolumn{2}{c}{Model}                                           & Mean AUC     & Most Improvement \\ \hline
		\multirow{3}{*}{GELN-freezed}   & \multicolumn{1}{c|}{Fusion model} & 85.1$\pm$0.0044 & 84.6                 \\
		& \multicolumn{1}{c|}{Graph model}  & 82.2$\pm$0.0093 & 82.5                 \\
		& \multicolumn{1}{c|}{GELN}         & 85.5$\pm$0.0038 & 85.3                 \\ \hline
		\multirow{3}{*}{GELN-umfreezed} & \multicolumn{1}{c|}{Fusion model} & 85.1$\pm$0.0093 & 84                   \\
		& \multicolumn{1}{c|}{Graph model}  & 81.6$\pm$0.0073 & 82.7                 \\
		& \multicolumn{1}{c|}{GELN}         & 85.9$\pm$0.0034 & 85.7                 \\ \hline
	\end{tabular}
	\label{table_4}
\end{table}

\begin{table}[]
	\centering
	\caption{The comparison using weighted average and mean average scheme in terms of AUC ($\%$).}
	\begin{tabular}{ccc}
		\hline
		Model                           & Scheme           & Mean AUC     \\ \hline
		\multirow{2}{*}{GELN-freezed}   & mean average     & 85.4$\pm$0.38 \\
		& weighted average & 85.5$\pm$0.38 \\
		\multirow{2}{*}{GELN-umfreezed} & mean average     & 85.8$\pm$0.33 \\
		& weighted average & 85.9$\pm$0.33 \\ \hline
	\end{tabular}
	\label{table_5}
\end{table}

\subsection{Ablation study}
We conduct the ablation study of our GELN-freezed and GELN-unfreezed and show their corresponding results, including the predictions from the fusion model, graph model, and final GELN, in Table~\ref{table_4}.
Both the graph model and fusion model are trained 50 times, and then we obtain the mean value and standard deviation to avoid the influence of random seed selection as much as possible and conduct a fair evaluation.
In Table~\ref{table_4}, it can be seen that for the GELN-freezed model, the mean AUC is improved from 85.1$\%$ of Fusion Model to 85.5$\%$ of GELN, while for the GELN-unfreezed model, the mean AUC is increased by 0.8$\%$ to $85.9\%$ compared to Fusion Model. 
We believe that it is because compared to freezing the fusion model, unfreezing the fusion model during the training of second step can learn more different information, which is complementary and helpful to the fusion model.
Also, we display the cases with the most improvement of AUC, among which the GELN-freezed can improve the performance of AUC from 84.6 $\%$ to 85.3$\%$ (0.7$\%$), and GELN-unfreezed can get significantly boost the performance from 84$\%$ to 85.7$\%$ (1.7$\%$). 
We also compare the results using the weighted average and average scheme in Table~\ref{table_5}.
From this table, we can see that both versions of GELN obtains better performance when using the weighted average scheme than the mean average scheme, therefore, the weighted average scheme is applied in our method.

\subsection{Evaluation in the ISIC 2016 dataset}
We also evaluate our GELN method in the other multi-label skin lesion classification dataset, i.e., ISIC 2016.
Note that we did not use the weighted average scheme when our GELN was applied in the ISIC-2016 dataset, as this dataset does not have the validation dataset.
Therefore, we use the mean average scheme, which sets the $W_{pf}$ and $W_{pg}$ as 0.5, where $W_{pf}$ and $W_{pg}$ are the weight of the $P_F$ and $P_G$ and $P_{total} = W_{pf}*P_F + W_{pg}*P_G$.
From Table~\ref{table_6}, we can see that compared to the Fusion model, our GELN ($W_{pf}$=0.5, $W_{pg}$=0.5) model also can improve the performance from 34.25$\%$ to 35.25$\%$.
We also compare two variants of GELN by setting $W_{pf}$=0.25, $W_{pg}$=0.75 and $W_{pf}$=0.75, $W_{pg}$=0.25.
These two variants also outperform the fusion model, while they are worse than GELN-unfreezed using the mean average scheme.


\begin{table}[t]
	\centering
	\caption{The performance comprison in the ISIC-2016 dataset in terms of AUC ($\%$).}
	\begin{tabular}{ccc}
		\hline
		\multicolumn{2}{c}{Model}                                      & Mean AUC(\%)     \\ \hline
		\multirow{5}{*}{GELN-unfreezed} & Fusion Model                 & 34.25$\pm$0.59 \\
		& Graph Model                  & 33.33$\pm$0.71 \\
		& GELN($W_{pf}$=0.5, $W_{pg}$=0.5)   & 35.25$\pm$0.39 \\
		& GELN($W_{pf}$=0.25, $W_{pg}$=0.75) & 35.20$\pm$0.39 \\
		& GELN($W_{pf}$=0.75, $W_{pg}$=0.25) & 34.80$\pm$0.48 \\ \hline
	\end{tabular}
	\label{table_6}
\end{table}

\section{Discussion}
\subsection{Comparison with previous methods}
One interesting phenomenon is that in Table~\ref{table_2}, our GELN method achieves the best AUC value and the balanced Sen and Spe values simultaneously.
For example, the difference of Sen and SPE values for our GELN-unfreezed is 19.7$\%$, but those values for Inception-combined, FM-FS, and GIIN are 21.7$\%$, 36.6$\%$, 19.8$\%$ and 27.9$\%$, respectively.
Also, the result of our GELN-unfreezed did not show an extremely difference between the Sen and Spe value, such as the AMFAM that obtains 0 Sen value and 100$\%$ Spe value for the case of VS-IR.
These results illustrate the superiority of our method in the dataset with an unbalanced class distribution, even if not using the class-balance schemes like Inception-combine used.

\subsection{Ablation study}
From Table~\ref{table_4}, we can see that GELN-unfreezed is better than GELN-freezed based on both Mean AUC and the case with the most improvement.
However, it can be known that in the second step, GELN-unfreezed does not use the pre-trained weights of the fusion model and freeze the fusion model. It requires an extra fusion model to keep the predictions only from CNNs. So. compared to GELN-freezed, GELN-unfreezed's parameters are almost two times that of GELN-freezed. 

\subsection{Evaluation in the ISIC 2016 dataset}
In Table~\ref{table_6}, we can observe that the Mean AUC of all the models is below 36$\%$.
It is because the number of lots classes in the ISIC-2016 dataset is zero, and thus its corresponding AUC is set to 0, such as the border classes for the categories of homogeneous area and streak.
This reason also contributes to lots of zero values in the diagonal of the correlation matrix in the ISIC-2016 dataset (See Fig.~\ref{fig_4}).

\subsection{Others}
From Table~\ref{table_4} and Table~\ref{table_6}, we can see that the GELN-unfreezed can achieve about 1$\%$ improvement of Mean AUC value, which illustrates that our method can be applied in different multi-label scenarios.
Furthermore, in Table~\ref{table_6},  GELN ($W_{pf}$=0.5, $W_{pg}$=0.5) and  GELN ($W_{pf}$=0.5, $W_{pg}$=0.5) achieve comparable performance and outperform  GELN ($W_{pf}$=0.25, $W_{pg}$=0.75), and in Table~\ref{table_4}, both GELNs using weighted average ($W_{pf}$=0.655, $W_{pg}$=0.455) achieves better performance than that using mean average ($W_{pf}$=0.5, $W_{pg}$=0.5), which illustrates that when we do not consider the prediction information from GCN as main details in the GELN but supplementary information of the prediction from fusion model,  the GELN can achieve the best performance.

Also, we try to replace GCN with graph attention network(GAT), but the result using GAT is much worse than GCN, so we select the GCN in our model.

\subsection{Future works}
There are two directions we want to study in the future.
On the one hand, a knowledge distillation scheme can be introduced to reduce the model's parameters of GELN-unfreezed without degrading the performance of this model or only sacrificing a few performance losses.
On the other hand, more relationships between diagnosis and visual characteristics from dermatologists' experiences can be added to the pipeline to improve performance.

\section{Conclusion}
We proposed a graph-ensemble learning network (GELN) to integrate the prior co-occurrence information into the pipeline, which was not fully explored by previous works.
To deal with the weak generalization ability problem of directly applying GCN,  GELN followed the idea of ensemble learning that considered the GCN as an auxiliary model to provide the complementary information to the fusion model and then selectively fuse the predictions from these two models to maximally utilize the valuable information and avoid the bad influence from GCN.
The experimental results show our GELN consistently improved the performance in different datasets and achieved the state-of-the-art performance in the well-recognized multi-label skin lesion dataset, i.e., the SPC dataset.

\end{document}